\documentclass{article}

\usepackage{PRIMEarxiv}

\usepackage[utf8]{inputenc} 
\usepackage[T1]{fontenc}    
\usepackage{hyperref}       
\usepackage{url}            
\usepackage{booktabs}       
\usepackage{amsfonts}       
\usepackage{nicefrac}       
\usepackage{microtype}      
\usepackage{lipsum}
\usepackage{fancyhdr}       
\usepackage{graphicx}       
\graphicspath{{media/}}     

\usepackage{cite}
\usepackage{amsmath,amssymb,amsfonts}
\usepackage{algorithmic}
\usepackage{graphicx}
\usepackage{textcomp}

\usepackage{bm}
\makeatletter
\AtBeginDocument{\DeclareMathVersion{bold}
\SetSymbolFont{operators}{bold}{T1}{times}{b}{n}
ˇ
\SetMathAlphabet{\mathrm}{bold}{T1}{times}{b}{n}
\SetMathAlphabet{\mathit}{bold}{T1}{times}{b}{it}
\SetMathAlphabet{\mathbf}{bold}{T1}{times}{b}{n}
\SetMathAlphabet{\mathtt}{bold}{OT1}{pcr}{b}{n}
\SetSymbolFont{symbols}{bold}{OMS}{cmsy}{b}{n}
\renewcommand\boldmath{\@nomath\boldmath\mathversion{bold}}}
\makeatother

\usepackage{amsmath,amsfonts}
\usepackage{algorithmic}
\usepackage{algorithm}
\usepackage{array}
\usepackage[caption=false,font=normalsize,labelfont=sf,textfont=sf]{subfig}
\usepackage{textcomp}
\usepackage{stfloats}
\usepackage{url}
\usepackage{verbatim}
\usepackage{graphicx}
\usepackage{cite}

\usepackage[utf8]{inputenc}

\usepackage{amsmath,amssymb}
\usepackage{amsthm}
\usepackage{enumerate}
\usepackage{caption}
\usepackage{float}

\usepackage{xcolor}  
\usepackage{acro}
\usepackage{mfirstuc}

\newcommand{\Lagr}{\mathcal{L}}

\newcommand{\is}[2]{#1 \in \textbf{#2}}

\theoremstyle{definition}
\newtheorem{definition}{Definition}[section]

\acsetup{first-style = long-short, list/display = used}

\DeclareAcronym{RL}{
  short=RL,
  long=Reinforcement learning,
}
\DeclareAcronym{MuJoCo}{
  short=MuJoCo,
  long=Multi-Joint dynamics with Contact,
}
\DeclareAcronym{GAN}{
  short=GAN,
  long=Generative Adversarial Network,
}
\DeclareAcronym{CycleGAN}{
  short=CycleGAN,
  long=Cycle-Consistent Generative Adversarial Network,
}
\DeclareAcronym{PlaNET}{
  short=PlaNET,
  long=Deep Planning Network,
}
\DeclareAcronym{UNIT}{
  short=UNIT,
  long=Unsupervised Image-to-Image Translation Networks
}
\DeclareAcronym{TL}{
  short=TL,
  long=Transfer learning,
}
\DeclareAcronym{MDP}{
  short=MDP,
  long=Markov decision process,
}

\DeclareAcronym{DM}{
  short=DM,
  long=decision making,
}

\title{Knowledge Transfer in Deep Reinforcement Learning via an RL-Specific GAN-Based Correspondence Function}

\title{Knowledge Transfer in Deep Reinforcement Learning via an RL-Specific GAN-Based Correspondence Function
\thanks{This work was supported by the join UTIA-PEF laboratory TALISMAN. The authors would like to acknowledge the contribution of the COST Action CA21169, supported by European Cooperation in Science and Technology.‌} 
}

\author{
  Marko Ruman \thanks{Department of Adaptive Systems, Institute of Information Theory and Automation, Czech Academy of Sciences} \\
  \texttt{ruman@utia.cas.cz} \\ 
   \And
  Tatiana V. Guy \footnotemark[2] \thanks{Department of Information Engineering, Faculty of Economics and Management, Czech University of Life Sciences} \\
  \texttt{guy@ieee.org} \\
}

\begin{document}

\maketitle

\begin{abstract}
Deep reinforcement learning has demonstrated superhuman performance in complex decision-making tasks, but it struggles with generalization and knowledge reuse—key aspects of true intelligence. This article introduces a novel approach that modifies Cycle Generative Adversarial Networks specifically for reinforcement learning, enabling effective one-to-one knowledge transfer between two tasks. Our method enhances the loss function with two new components: model loss, which captures dynamic relationships between source and target tasks, and Q-loss, which identifies states significantly influencing the target decision policy.

Tested on the 2-D Atari game Pong, our method achieved 100\% knowledge transfer in identical tasks and either 100\% knowledge transfer or a 30\% reduction in training time for a rotated task, depending on the network architecture. In contrast, using standard Generative Adversarial Networks or Cycle Generative Adversarial Networks led to worse performance than training from scratch in the majority of cases. The results demonstrate that the proposed method ensured enhanced knowledge generalization in deep reinforcement learning.

\end{abstract}


\noindent\begin{minipage}{0.5\textwidth}
 \section*{Notation used}
\begin{tabular}{p{0.09\textwidth}p{0.82\textwidth}}
    $s_t$ & State at the $t$-th time step \\
    $r_t$ & Reward received at the $t$-th time step \\
    $a_t$ & Action chosen at the $t$-th time step\\
    $\gamma$ & Discount factor\\
    $R$ & Reward function \\
    $Q$ & $Q$-function \\
    $\mathcal{C}$ & Correspondence function \\
    $F$ & Environment model \\
    $\textbf{K}$ & Knowledge gained from a task \\
    $\textbf{M}$ & Experience memory \\    
    $G_S$ & Generator from source to target task\\
    $G_T$ & Generator from target to source task\\
    $\Lagr_{GAN}$ & GAN loss\\
	$\Lagr_{Cyc}$ & Cycle-consistency loss\\
	$\Lagr_{Q}$ & $Q$-loss\\
	$\Lagr_{M}$ & Model loss\\
    $\lambda_{Cyc}$ & Weight of cycle-consistency loss\\

    $\lambda_{Q}$ & Weight of $Q$-loss\\
    $\lambda_{M}$ & Weight of model loss\\
\end{tabular}

\end{minipage}
\begin{minipage}{0.5\textwidth}
 \section*{Acronyms}
\vspace{0.2cm}
\begin{tabular}{p{0.20\textwidth}p{0.70\textwidth}}
	$DM$ & Decision making \\
    $GAN$ & Generative Adversarial Network \\
    $CycleGAN$ & Cycle-Consistent GAN \\
    $MDP$ & Markov decision process\\
    $MuJoCo$ & Multi-Joint dynamics with Contact\\
    $PlaNET$ & Deep Planning Network\\
    $RL$ & Reinforcement learning\\
    $TL$ & Transfer learning\\
    $UNIT$ & Unsupervised Image-to-Image Translation Networks
\end{tabular}
\end{minipage}

\section{Introduction}

The inherent ability of \ac{RL} to dynamically learn complex policies through trial and error has shown great potential in solving diverse decision problems. Deep \ac{RL}, which combines the advantages of \ac{RL} with the power to handle high-dimensional data, has recently brought many advances. 
For instance, model-free methods have shown significant results in \ac{MuJoCo} environments \cite{mujoco}, real-world robotic applications \cite{realRL}, and have demonstrated an ability to achieve super-human performance in Atari games \cite{dqn}, \cite{advanceddqn}. Model-based deep \ac{RL} methods such as AlphaZero \cite{alphazero} and \ac{PlaNET} \cite{planet} have also made significant progress. However, \ac{RL} remains unsuitable for many real-world tasks, as errors can be extremely costly. One promising way to address this issue is through \emph{\ac{TL}}  \cite{tlsurvey}, \cite{Zhu_etal:23}, where skills and knowledge collected from similar tasks are applied to the current problem. 
Additionally, \ac{TL} plays a crucial role in: developing agents capable of lifelong learning \cite{lifelongrl}, multi-task learning \cite{Varghese:21}, enabling simulation-to-real knowledge transfer in robotics \cite{sim2realgrasp, retinagan, rlcyclegan, sim2realsurvey, Ranaweera:23}, and advancing the development of general AI \cite{Silver_etal:21}, \cite{generalai}.

Despite many advances, the use of transfer learning in \ac{RL}, especially in deep RL, remains limited due to several challenges:
\begin{itemize}
	\item \textit{Weak ability of the \ac{RL} agent to generalize to unobserved tasks}. For example, the output of a deep convolutional network for image data can be dramatically altered by a $1$-pixel perturbation of the input image, \cite{onepixelworl}. This issue extends to \ac{RL}, as image data often form the observable states in \ac{RL} tasks. For instance, $1$-pixel perturbations can lead to ineffective policies, \cite{onepixelrl}. \ac{RL} methods frequently fail to reuse previously acquired knowledge even in similar tasks when the original image is rotated or when some colours are changed. It has also been shown that learning from scratch can be more efficient than fine-tuning a previously obtained model \cite{Gamrian}. This significantly contrasts with the human ability to generalize and reuse previously acquired knowledge.
	\item \textit{Challenging transfer of knowledge and experience} from previously solved tasks to unseen ones. A decision policy learnt from similar tasks may not always be effective in solving the current decision task. For example, optimal policies for driving a motorcycle in racing conditions are unsuitable and even dangerous for public roads.
\end{itemize}

The objective of this paper is to create an efficient method for one-to-one knowledge transfer between different \ac{RL} tasks, with the aim of \emph{improving} the agent's performance on the target task and \emph{reducing} the training time required. The primary motivation is that any transferred object (skills or knowledge) is typically task- and policy-specific\footnote{The environment dynamics and rewards of \ac{RL} tasks may differ.}. To ensure that the transferred skills are relevant and effective, we focus on identifying and transferring the most informative patterns. This approach enhances the \ac{RL} agent's ability to generalize across tasks, thereby improving performance in the target task.

An additional, but important, reason why \ac{TL} may fail is that the tasks involved can differ significantly in their dynamics and rewards. The proposed method addresses this by considering the matching of the dynamics of the involved \ac{RL} tasks to ensure that appropriate behaviour patterns are transferred.

The proposed solution adopts a cyclic paradigm and is formulated as an \emph{\ac{RL}-specific modification of CycleGAN}. It introduces two new components to the loss function: model loss and $Q$-loss. The model loss captures the essential dynamic relationship between the involved \ac{RL} tasks, while $Q$-loss prioritises states that affect learning the policy of the target \ac{RL} task.

\subsection*{Main contributions of the paper}

\begin{itemize} 
\item [$\bullet$] \textbf{Introduction of an efficient method for knowledge transfer.} We propose a novel method for knowledge transfer between two different \ac{RL} tasks based on an \ac{RL}-specific modification of \ac{CycleGAN}, designed to enhance generalization and reduce training time across tasks.

\item [$\bullet$] \textbf{Establishment of a correspondence function.} We develop a correspondence function that learns and reveals the similarities between source and target \ac{RL} tasks. This function plays a key role in facilitating efficient and accurate knowledge transfer.

\item [$\bullet$] \textbf{Development of a four-component loss function.} Our four-component loss function incorporates model loss, $Q$-loss, and two additional components to better reflect task dynamics and account for the actual policy being used. This design improves the transfer of relevant knowledge and enhances the learning process.

\item [$\bullet$] \textbf{Generalization of \ac{GAN} and \ac{CycleGAN} methods.} By introducing two new components to the loss function, we extend and generalize the \ac{GAN} and \ac{CycleGAN} methods. Our approach demonstrates that these standard techniques are special cases of our proposed framework, providing a broader, more powerful solution for knowledge transfer in \ac{RL}.

\item [$\bullet$] \textbf{Complete knowledge reuse in Pong tasks.} We achieve 100\% knowledge reuse in experiments transferring between the original Pong and a \emph{rotated} Pong environment, highlighting the method's ability to fully transfer learned skills without requiring re-training.

\item [$\bullet$] \textbf{Handling of challenging tasks where standard methods fail.} Our method successfully handles tasks that are problematic for traditional \ac{GAN} and \ac{CycleGAN} methods, allowing for faster learning and improved performance where the only alternative would be to learn from scratch.

\end{itemize}

\subsubsection*{Advantages of the proposed method:}

\begin{itemize}

   \item \textbf{No reliance on paired data}: Unlike many transfer learning techniques, our approach does not require paired datasets, making it broadly applicable across various domains and tasks.
   \item \textbf{Task and domain independence}: The method is independent of the nature of the \ac{RL} tasks involved, meaning it can be applied to diverse domains without task-specific manual engineering.
   \item \textbf{Flexible data formats}: It works with different data formats for states (e.g., images, sounds, numerical vectors), ensuring versatility and applicability to a wide range of applications. The method can be adapted to various RL tasks by selecting an appropriate network architecture tailored to the specific data format, not limited to image data alone.
   
\end{itemize}
 
 The paper layout is as follows. Section~\ref{ch:Q} recalls the necessary background and formulates the considered \ac{TL} problem. Section~\ref{ch:tl learning} constructs the correspondence function and proposes a novel method of its learning. Section~\ref{ch:experiment} describes the experimental evaluation of the proposed approach and compares it with baseline methods. Section~\ref{ch:conclusion} provides concluding remarks and outlines future research directions.

\subsection*{Related works}

Survey \cite{Zhu_etal:23} systematically analyses recent advances in transfer learning for deep \ac{RL}. Our research approach falls within the category of methods that employ mapping functions between the source and target tasks to facilitate knowledge transfer. A notable subset of this research focuses on learning shared features across \ac{RL} tasks that are transferable. As demonstrated in \cite{midlevelfeatures}, policies trained on intermediate-level features, referred to as \emph{mid-level features}, exhibit superior generalization compared to policies trained directly on raw image observations. Work~\cite{invariantfeaturesrl} leverages general features of two \ac{RL} tasks with different dynamics. However, the method is based on paired image observations which are hard or impossible to obtain in practice. Work \cite{barreto} achieved success in tasks differing in reward function by maintaining successor features and decoupling environment dynamic and reward function. Approach \cite{knowledgeshapingtl} introduces task similarity criterion and builds \ac{TL} framework based on knowledge shaping, where for similar tasks, efficient transfer is theoretically guaranteed.   \\

 The pioneering work that used task correspondence was based on unsupervised image-to-image translation models \ac{CycleGAN}, \cite{cyclegan}, and \ac{UNIT}, \cite{unit}. Approach \cite{Gamrian} achieved results on a specific set of tasks by finding correspondence between states of two \ac{RL} tasks. The application potential of the approach is rather limited as problems like mode-collapse are present. Works \cite{rlcyclegan} and \cite{retinagan} improved the approach proposed in \cite{Gamrian} by introducing $Q$-function or object detection into the learning of the task correspondence. One of the recent approaches, \cite{dynamiccycle}, considers an environment model while learning the task correspondence, which is strongly inspired by the video-to-video translation model, \cite{videogan}.

\section{Background and notation}
\label{ch:Q}
This section briefly recalls \ac{RL} formalism and introduces the considered problem. 
\subsection{Notation}
\label{s:notation}
\noindent Throughout the text, sets are denoted by bold capital letters (e.g. $\textbf{X}$), $\mathbb{N}$ and $\mathbb{R}$ are sets of natural and real numbers respectively. 
$\left\lVert x\right\lVert$ is the L1 norm of $x$. $x_t$ is the value of $x$ at discrete time $t \in {\mathbb{N}}$. $E_p[x]$ denotes the expected value of $x$ with respect to a probability density $p$ (if provided). Specific notations are provided at the beginning of the article.

	We formalise the transfer problem in a general way by considering 
	 two \ac{RL} tasks - the \emph{source task}, $S$, and the \emph{target task}, $T$, characterised by their respective task domains. $\textbf{S}_S\times\textbf{A}_S$ and $\textbf{S}_T\times\textbf{A}_T$, with $\textbf{S}$ and $\textbf{A}$ denoting a set of states and a set of actions respectively.	 

\subsection{Reinforcement learning}
\label{s:definition}

Reinforcement learning (\ac{RL}) considers an \emph{agent} purposefully interacting with an \emph{environment} by selecting actions. \ac{RL} agent models its environment as \emph{\ac{MDP}}, \cite{puterman} consisting of discrete sets of observable states $\textbf{S}$ and actions $\textbf{A}$. Set $\textbf{S}\times\textbf{A}$ is referred to as the \emph{task domain}. At each time $t$, the agent observes environment state $\is{s_t}{S}$ and takes action $\is{a_t}{A}$. Executing action ${a_t}$ at state ${s_t}$: i) causes a transition of the environment to state $s_{t+1}$ according to \emph{transition function} that describes $p(s_{t+1}| s_t, a_t)$, and ii) provides reward $r_t$, i.~e. the value of reward function $R(s_{t+1}, a_t, s_t): \textbf{S} \times \textbf{A} \times \textbf{S} \mapsto \mathbb{R}$. The agent's \emph{goal} is to learn policy $\pi: \textbf{S} \mapsto \textbf{A} $ that maximises the accumulated reward. 
																																																	
The solution of a \ac{MDP} task is the \emph{optimal} policy $\pi^*$:
\begin{equation}\label{e:main task}
		\pi^* = \underset{\pi\in\boldsymbol{\Pi}}{\mathrm{\arg \max}}\hspace{0.07cm}E\left[\sum_{t=1}^N\gamma^tR(s_{t+1}, \pi(s_{t}), s_t)\right]
        \text{where} \boldsymbol{\Pi}=\left\{\pi(s_{t})\right\}_{t=1}^{N}
\end{equation}
with \textit{decision horizon} $N \in \mathbb{N}\cup \{\infty\}$ and \textit{discount factor} $\gamma \in (0,1)$.
The \ac{RL} agent learns to act optimally within \ac{MDP} when the transition function and reward function are unknown. A good \ac{RL} modifies $\pi$ over time to gradually get it closer to an optimal policy.  
$Q$-learning, a model-free \ac{RL} algorithm, is one of the traditional solution approaches. It aims to learn $Q$-function (aka \emph{state-action-value function}) that quantifies the expected value of future discounted reward over the states induced by $\pi^*$ for given starting state $s$ and action $a$. 
\begin{equation}
\hspace{-5pt}	Q(s,a) = E_{\pi^*}\left[\sum_{t=1}^N\gamma^t R_{t}(s_{t+1}, \pi^*(s_t), s_t) \Big{|} s_1=s, a_1=a\right].
	\label{e:q-function}
\end{equation}
Discount factor $\gamma$ expresses the agent's preferences towards immediate reward over future ones.
The estimate of (\ref{e:q-function}), $\hat{Q}(s,a)$, can be gradually learnt on the stream of data records $(s_t, a_t, r_t, s_{t+1})$ using for instance, temporal difference learning, \cite{qlearning}:
\begin{equation}
	\hat{Q}_{t+1}(s_t,a_t) = (1-\alpha)\hat{Q}_t(s_t,a_t) + \alpha(r_t + \gamma \hspace{0.07cm}\underset{\is{a}{a}}{\mathrm{max}}\hspace{0.07cm}\hat{Q}_t(s_{t+1},a)),
\end{equation}
where $\alpha \in (0,1)$ is a parameter called \emph{learning rate} and $r_t = R(s_{t+1}, a_t, s_t)$. The learning starts with an initial estimate of the $Q$-function, $\hat{Q}_0(s,a)$. 
The learned, approximately optimal, decision rule is then
\begin{equation}
	\pi^*\left(s\mid\hat{Q}\right) = \underset{\is{a}{a}}{\mathrm{argmax}}\hspace{0.07cm} \hat{Q}(s,a).
\end{equation}

\subsection{Deep Q-learning}
\label{s:dqn}

Whenever the state space is huge, for instance, when the state is given by a video frame, efficient learning of $Q$-function calls for numerical approximation. The state-of-the-art in function approximation points to deep neural networks (DNN) as a suitable methodology, \cite{nn}. 

Deep $Q$-networks (DQN), \cite{mnih2015human}, use a standard off-policy $Q$-learning, \cite{qlearning}, and DNN to estimate the $Q$-function (\ref{e:q-function}).

DQN approximates $Q$-function by a deep neural network with parameters that can be trained similarly to the supervised learning, \cite{dqn}. However, the supervised learning assumes i.~i.~d.\ input data. Moreover, output values are expected to be the same for the same inputs, \cite{deepsupervisedlearning}. Neither of these assumptions is met in \ac{RL} tasks. The consecutive states are usually highly correlated (e.~g.~video frames) and thus very far from being i.~i.~d.\ Output values also contain learned $Q$-function that evolves during learning. This makes the learning process unstable.
To enable data reuse and stabilise the learning, DQN uses an \textit{experience replay technique} to remove correlations in the observed sequence and employs an additional \textit{target network}\footnote{Note that name \emph{target network} in DQN generally does not refer to \emph{target task}.} to stabilise the output values, see \cite{dqn} for details.

\textbf{Experience replay technique} considers that the last $n_M$ data records (so-called \emph{experience memory}, denoted as $\textbf{M}$) are stored in a memory buffer. At each learning step, a mini-batch of length $n_B \in \mathbb{N}$ is randomly sampled from the memory buffer and is used to update the neural network that approximates $Q$-function. It brings the learning data closer to being i.~i.~d.\

\textbf{Target network} is an additional network\footnote{that has the same architecture as the original network} serving for stabilising the learning. The idea is as follows. The parameters of the \emph{original network} are updated at every learning step, while \emph{target network} is used to retrieve output values and stays static, i.e. its parameters do not change. Every $n_U \in \mathbb{N}$ steps, the original and target networks are synchronised.
Details on the DQN algorithm, see Appendix.

\subsection{Cycle-Consistent GAN}
\label{s:CycleGAN}
\ac{CycleGAN}, \cite{cyclegan}, is based on \ac{GAN}, \cite{gan}, and was originally proposed for image-to-image translation. The idea behind \emph{cycle consistency} is that data that has been translated to a new domain and then recovered from it, should not change. 

\ac{CycleGAN} operates with two mappings $G_S$ and $G_T$ called \emph{generators}\footnote{that translate data between source and target domains} 
\begin{equation}
	\label{e: generators}
	G_S: \hspace{0.1cm} \textbf{S}_S \rightarrow \textbf{S}_T \hspace{0.3cm} \text{and} \hspace{0.3cm}  G_T: \hspace{0.1cm}\textbf{S}_T \rightarrow \textbf{S}_S.
\end{equation}
They are learnt as two \ac{GAN}s, that is, simultaneously with the corresponding discriminators $D_S$ and $D_T$. Generators learn to map states from $\textbf{S}_S$ to $\textbf{S}_T$ and vice-versa, while discriminators learn to \emph{distinguish} a real state from a state mapped by a generator.
Mappings $G_S$, $G_T$, $D_S$ and $D_T$ are constructed as neural networks with their architecture depending on the data format. For instance if states are images, convolutional layers are often used. 

Learning in \ac{CycleGAN} minimises a two-component loss. The first is \emph{adversarial loss}, $\Lagr_{GAN}$ comes from \ac{GAN} and is given by
\begin{align}
	&\Lagr_{GAN} = \hspace{0.03cm} 
	 E_{s_S}\left[\mathrm{log}\hspace{0.03cm}D_S(s_S)\right] + E_{s_T}\left[\mathrm{log}\left(1-D_S\left(G_T\left(s_T\right)\right)\right)\right]
	 + E_{s_T}\left[\mathrm{log}\hspace{0.03cm}D_T(s_T)\right] + E_{s_S}\left[\mathrm{log}\left(1-D_T\left(G_S\left(s_S\right)\right)\right)\right]
	\label{e:gan_loss}
\end{align}
The adversarial training encourages mappings $G_S$ and $G_T$ (\ref{e: generators}) to produce outputs indistinguishable from the real ones, i.~e. respective sets $\textbf{S}_S$ and $\textbf{S}_T$. However, minimising $\Lagr_{GAN}$ does not prevent the network from mapping the same set of input images to any permutation of images in the target domain.

The second component is \emph{cycle-consistency} loss, $\Lagr_{Cyc}$, that has the following form:
\begin{align}
	\label{e:cycle_loss}
	\Lagr_{Cyc}  = \hspace{0.03cm} E_{s_S}\left[ \left\lVert G_T\left(G_S\left(s_S\right)\right) - s_S\right\rVert\right] 
	+ E_{s_T}\left[\left\lVert G_S\left(G_T\left(s_T\right)\right) - s_T\right\rVert\right].
\end{align}
Minimisation of cycle-consistency loss $\Lagr_{Cyc}$ ensures that every state $s_S \in \textbf{S}_S$ must be recoverable after mapping it back to $\textbf{S}_T$, i.e. $G_T(G_S(s_S)) \approx s_S$. The same requirement applies to every state $s_T \in  \textbf{S}_T$.

\section{Transfer learning for \ac{RL}}
\label{ch:tl learning}
Humans have a remarkable ability to generalise. They do not learn everything from scratch but rather reuse earlier acquired knowledge to a new task or domain\footnote{Developmental psychologists have shown that as early as 18 months old, children can infer intentions and imitate the behaviour of adults, \cite{psychology}. The imitation is complex as children must infer a match between their observations and internal representations, effectively linking the two diverse domains.}. Generally, finding common patterns between different tasks and effectively transferring the concepts learned from one task to another is an essential characteristic of high-level intelligence. Thus, the efficient solution of transfer learning will allow for the creation of intelligent agents that can mimic human thinking and solve problems in a much more explainable way. Moreover, efficient reusing the acquired knowledge may accelerate the learning process and make complex tasks learnable. 

This section, we formalises a problem of transfer learning between two \ac{RL} tasks, empirically introduces a correspondence function reflecting the similarity of two \ac{RL} tasks and proposes an \ac{RL}-specific modification of \ac{CycleGAN} algorithm that realises knowledge transfer between two \ac{RL} tasks. The proposed transfer i) considers behaviours, which are most useful for the target task; ii) captures and respects common patterns in transition dynamics of the involved \ac{RL} tasks. 

\subsection{Problem formulation}

  We consider two \ac{RL} tasks: the \emph{source task}, $S$, and the \emph{target task}, $T$ with their respective task domains $\textbf{S}_S\times\textbf{A}_S$ and $\textbf{S}_T\times\textbf{A}_T$. Each of the tasks corresponds to \ac{MDP} with its own environmental dynamics and reward function, see Section~\ref{s:definition}. Transition functions of the tasks as well as theirs reward functions may be different. 
  
  Intuitively, the success of transfer between two \ac{RL} tasks depends on the degree of similarity between these tasks. If the tasks are dissimilar, the transfer of inappropriate knowledge may significantly worsen the resulting performance in the target task. Therefore, the success of the transfer broadly depends on the existence of some common properties between the source and target tasks. The similarity can be perceived from various perspectives, such as sharing the same environment, obeying similar laws of physics, or involving similar objects for interaction. For instance, when driving a motorcycle, encountering an animal on the road may correspond to pulling the brake levers, just as when driving a car, the sight of a person crossing the road can lead to pressing the brake pedal.
  
  This work uses an abstract notion of similarity, inspired by human learning when tackling related problems. Two tasks are similar if they share some common properties, and the knowledge acquired in one task proves to be beneficial in solving the other. This empirical definition can be more formally introduced as follows.
  
\begin{definition} [Correspondence function]
	\label{def:task similarity}
	Consider source $S$ and target $T$ tasks with respective domains $\textbf{S}_S\times\textbf{A}_S$ and $\textbf{S}_T\times\textbf{A}_T$. 
	A \emph{correspondence function}, $\mathcal{C}: (\textbf{S}_T\times\textbf{A}_T) \mapsto (\textbf{S}_S\times\textbf{A}_S)$, is a mapping, which reveals the similarity of the involved \ac{RL} tasks in terms of the dynamics of the tasks' environments and the associated $Q$-functions.

\end{definition}

It is clear that function $\mathcal{C}$ establishes the relationship between similar patterns in behaviour of the target and source tasks that are necessary for knowledge transfer. So, if $Q_S$ is the optimal $Q$-function for the source task, then $Q$-function 
	\begin{equation}
	Q_S(\mathcal{C}(.,.)) : \textbf{S}_T\times\textbf{A}_T \mapsto \mathrm{R}
\end{equation}
gives better performance\footnote{Performance is measured by average reward per time.} on the target task than a random policy. 

Let us assume (for brevity) that the action spaces of the source and the target \ac{RL} task are identical, i.e. $\textbf{A}_S = \textbf{A}_T$. Let mutually corresponding actions be found using identity mapping regardless of the current state\footnote{More specifically, all actions of the source and target task have the same labels and meanings (e.g. $a=1$ stands for "up"). Therefore, no mapping between source and target task action spaces is necessary}. Thus, we need to learn a mapping indicating corresponding states, i.~e. the correspondence function for states.
The searched correspondence function $\mathcal{C}$  is then obtained as follows:
\begin{equation}
	\label{e: specific correspondence function}
	\mathcal{C}(s_T, a_T) = (G_T(s_T), I(a_T)), \hspace{0.5cm} \forall (s_T,a_T) \in \textbf{S}_T\times\textbf{A}_T,
\end{equation}
where $G_T$ is the generator from (\ref{e: generators}) mapping states from the \emph{target task} to states from the \emph{source task} and $I(.)$ is an identity mapping.

The correspondence function is unknown to \ac{RL} agent and the next section describes how to learn it.

\subsection{Learning of correspondence function}
\label{s:correspondence learning}
 
The proposed learning is inspired by \ac{CycleGAN}, see Section~\ref{s:CycleGAN}, where the learning minimises a discriminative \emph{loss function}, which makes the similarity metric small for similar patterns and large otherwise. 
Even direct application of \ac{CycleGAN} to the states brought some success in policy transfer, see for instance \cite{Gamrian}. However, data records in experience memories comprise richer yet unused information that may be helpful for the transfer of knowledge. 
We propose to include additional components into the loss function minimised in \ac{CycleGAN} learning. They will consider unused information and make the learned correspondence entirely relevant to \ac{RL}. 
The proposed loss will ensure that the learnt function $\mathcal{C}$ captures all patterns significant for the intended \ac{TL}. In particular, the loss should respect both the dynamics and $Q$-function of the source task.

This work proposes adding two new components to the \ac{CycleGAN} losses, \eqref{e:gan_loss}, \eqref{e:cycle_loss}:
\begin{itemize}
	\item $Q$-loss $\Lagr_Q$ - a loss that reflects how the $Q$-function learned from the source task, $Q_S$, 
	 copes with impreciseness in learned generators $G_T$ and $G_S$. 
	\item Model-loss $\Lagr_M$ - a loss that reflects the influence of the environment model of the source task.
\end{itemize}
Let us explain the reasons for introducing the new components and their forms.
\subsubsection*{$\boldsymbol{Q}-${\bf{\textit{loss}}}}

The $Q$-function, $Q_S$, plays a central role in \ac{RL} as it defines the optimal policy for the source task. When transferring knowledge from a source task $S$ to a target task $T$, it is essential to preserve the $Q$-function’s accuracy for states relevant to the \ac{DM} process. This motivates the introduction of $Q$-loss, $\Lagr_Q$, which ensures that the learned correspondence function, $\mathcal{C}$, maintains consistency between the value estimates of corresponding states in both domains.

The \emph{cycle-consistency} loss (\ref{e:cycle_loss}) ensures that the generators $G_S$ and $G_T$ map between the source and target domains in a consistent way. However, cycle-consistency alone does not prioritize the states that are most critical for decision-making in \ac{RL}. The $Q$-loss directly incorporates the $Q$-function, encouraging the generators to focus on the states that matter most for choosing optimal actions. Mathematically, the $Q$-loss is defined as: \begin{align} \label{e:q_loss}
 \Lagr_{Q} = \hspace{0.03cm} &E_{a,s_S}\left[ \left\lVert Q_S\left(G_T\left(G_S\left(s_S\right)\right), a\right) - Q_S\left(s_S, a\right)\right\rVert\right] \end{align}

This loss minimizes the difference between the $Q$-function values of state $s_S$ in the source task and its mapped counterpart after a round-trip through the generators ($G_T(G_S(s_S))$). In simpler terms, this forces the correspondence function $\mathcal{C}$ to retain the critical information from the states in the source domain that is essential for determining the optimal policy, ensuring that this information is preserved after mapping between tasks $S$ and $T$.

The rationale behind this is that for effective knowledge transfer in \ac{RL}, it is not enough for the state representations to be similar visually or structurally; they must also be similar in terms of their impact on decision-making, as captured by the $Q$-function. By focusing on states that are important for action selection, the $Q$-loss makes the correspondence function more suitable for transferring policies between tasks.

\subsubsection*{\bf{\textit{Model loss}}}
In reinforcement learning, tasks are inherently dynamic, meaning that a state’s importance often depends on the actions taken and how the state evolves over time. This dynamic nature introduces a key challenge for transferring knowledge between tasks, as it is not just the individual states that matter but the transitions between them. To address this, we introduce the \emph{model loss}, $\Lagr_M$, which ensures that the correspondence function respects the underlying dynamics of the source and target tasks.

While losses like $\Lagr_{GAN}$, $\Lagr_{Cyc}$, and $\Lagr_Q$ focus on individual state mappings, they do not ensure that the temporal dynamics of the target task align with those of the source task. In other words, even if individual states match, the transitions between states (due to actions taken) might not be consistent. The model loss addresses this by incorporating the environment model $F_S$ of the source task. The model $F_S$ predicts the next state based on the current state and action: \begin{align} F_S: (s_t, a_t) \mapsto s_{t+1} \end{align}

To ensure that the learned correspondence function, $\mathcal{C}$, captures the dynamic relationships between the source and target tasks, we define the model loss $\Lagr_M$ as: \begin{align} \label{e:model_loss}
 \Lagr_{M} = \hspace{0.03cm} &E_{s_{Tt},a_{Tt},s_{Tt+1}}\left[ \left\lVert F_S\left(G_T\left(s_{Tt}\right), a_{Tt}\right) - G_T\left(s_{Tt+1}\right) \right\rVert\right] \end{align}

This loss ensures that the transitions in the target task are consistent with those in the source task when mapped through the correspondence function. Specifically, if an action $a_{Tt}$ taken in the target task leads to a state transition from $s_{Tt}$ to $s_{Tt+1}$, the model loss ensures that this transition corresponds to a valid transition in the source task. In other words, applying $a_{Tt}$ to the mapped state $G_T(s_{Tt})$ should lead to a state $G_T(s_{Tt+1})$ that is predicted by the source task’s environment model $F_S$.

Intuitively, this means that the correspondence function not only matches individual states between the source and target tasks but also ensures that the way states evolve over time (due to actions) is consistent. This is crucial for transferring knowledge about dynamic tasks, where the sequence of states and actions is key to solving the problem.

In summary, the model loss ensures that the correspondence function respects the temporal dynamics of both the source and target tasks, making it suitable for transferring policies between dynamic \ac{RL} tasks. Together, the $Q$-loss and model loss guarantee that the transferred knowledge is useful both for individual states and for the dynamic relationships between them.

\subsubsection*{\bf{\textit{Total loss}}}
\label{s:final_loss}
The proposed total loss comprises all the components (\ref{e:gan_loss}), (\ref{e:cycle_loss}), (\ref{e:q_loss}) and (\ref{e:model_loss}) and, thus, has the following form:
\begin{align}
	\label{e:final loss}
	\Lagr  = \hspace{0.03cm} \Lagr_{GAN} + \lambda_{Cyc}\Lagr_{Cyc} + \lambda_{Q}\Lagr_Q + \lambda_M\Lagr_M,
\end{align}
where $\lambda_{Cyc}$, $\lambda_{Q}$ and $\lambda_M$ are \emph{loss parameters} that define relative influence (weight) of the respective components.

The proposed approach, which minimises $4$-component loss \eqref{e:final loss}, generalises \ac{GAN}, \cite{gan}, and \ac{CycleGAN}, \cite{cyclegan}, methods often used for transfer learning. It is easy to see that \ac{GAN} and \ac{CycleGAN} can be obtained by setting some of parameters $\lambda_{Q}$, $\lambda_M$, $\lambda_{Cyc}$ in (\ref{e:final loss}) to zeros as follows:
\begin{itemize}
	\item $\lambda_Q =  \lambda_M = \lambda_{Cyc} = 0$ (for \ac{GAN}),
	\item  $\lambda_Q =  \lambda_M = 0$ (for \ac{CycleGAN}).
\end{itemize}

\subsection{Transfer learning: Algorithm}
\begin{figure}[H]
	\centering
	\includegraphics[width=.5\linewidth]{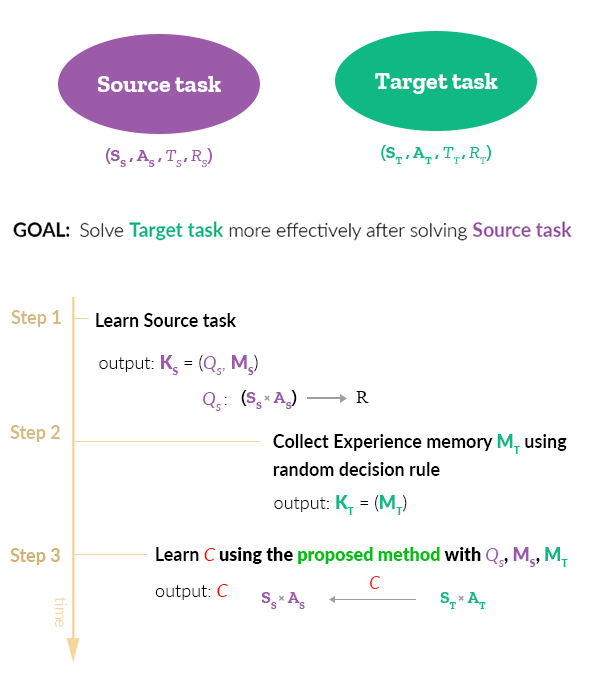}
	
	\caption{\textbf{The proposed \ac{TL} between tasks $S$ and $T$}.\label{figure:TL}}	
\end{figure}
The main steps of the proposed algorithm:
\begin{description}
	\item [\textit{Step 1}] The agent first solves task $S$ by the DQN algorithm (see Section \ref{s:dqn}). The obtained knowledge, $\textbf{K}_S = (Q_S, \textbf{M}_S)$, consists of learned $Q$-function, $Q_S$, and collected experience memory $\textbf{M}_S = \left(\left(s_t, a_t, s_{t+1}, r_t\right)_{i=1}^{n_M}\right)$. 
	\item [\textit{Step 2}] The agent applies a random decision rule to task $T$, collects experience memory $\textbf{M}_T$. Further the agent uses $\textbf{M}_T$ together with knowledge $\textbf{K}_S$, to solve \emph{target task} $T$ more efficiently, see Figure \ref{figure:TL}.
	\item [\textit{Step 3}] The assumed similarity of the tasks $S$ and $T$ guarantees the existence of correspondence function $\mathcal{C}$ (see Definition~\ref{def:task similarity}). The agent uses knowledge $\textbf{K}_S = (Q_S, \textbf{M}_S)$ and memory $\textbf{M}_T$ to learn correspondence function $\mathcal{C}$. Hence the correspondence function is used to transform state-action pairs from the target task to the source task.
	\item [\textit{Step 4}] Existence of a correspondence function $\mathcal{C}$, allows to express $Q$-function of the target task, $Q_T$, via $Q$-function of the source task, $Q_S$, and learnt correspondence function $\mathcal{C}$ as follows:
	\begin{equation}
		\label{e: transformed Q}
		\hspace{-0.8cm} Q_T(s_T, a_T) = Q_S(\mathcal{C}(s_T, a_T)), \hspace{0.1cm} \forall (s_T,a_T) \in  \left(\textbf{S}_T\times\textbf{A}_T\right).
	\end{equation}
\end{description}

 Then the agent can use $Q$-function $Q_S$ of the source task to choose the optimal actions in the target task. 

\subsubsection*{Note on implementation} Similarly to \ac{GAN}, in the considered case of \ac{TL} we have two experience memories with \emph{mutually unpaired} entries: $\textbf{M}_S$ for the source task and $\textbf{M}_T$ for the target task. The proposed algorithm  learns the correspondence function, $\mathcal{C}$, that will match them.

The experience memory $\textbf{M}_S$ is obtained as a by-product of DQN algorithm used for learning the optimal $Q$-function $Q_S$. However, the proposed method does not strictly require usage of DQN. It is important that it can be applied to any algorithm giving $\textbf{M}_S$ and $Q_S$ (where $Q_S$ is a differentiable function).

\subsubsection*{Note on the use}
This paper considers using the transfer learning method just once, in the beginning of interaction with the target task. Other ways, however, might be explored such as when partially optimal strategy is found to further improve it.

\section{Experimental part}
\label{ch:experiment}

To test the efficiency of the proposed approach, two experiments on the Atari game Pong, \cite{atari}, were conducted. The performance of the approach was evaluated based on an average accumulated reward per game. \ac{GAN} and \ac{CycleGAN} were used as baseline methods.

\subsection{Domain description}

Pong is a two-dimensional game simulating table tennis. There are six available actions ('do nothing', 'fire', 'move up', 'move down', 'move up fast', 'move down fast'). The last four observed image frames served as a task state. The agent learned to play the game using the DQN algorithm, Section \ref{s:dqn}, and, thus, learned the $Q$-function. To test the approach described in Section \ref{s:correspondence learning}, the agent also learned environment model $F$.

\subsection{Experiment description and setup}
The proposed \ac{TL} method was tested in two experiments.

\begin{figure*}[!t]
\centering
\begin{minipage}{.5\textwidth}
  \centering
  \includegraphics[width=.4\linewidth]{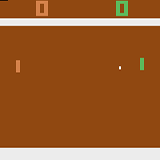}
  \captionof{figure}{\textbf{Standard Pong}, \cite{atari}}
  \label{fig:test1}
\end{minipage}%
\begin{minipage}{.5\textwidth}
  \centering
  \includegraphics[width=.4\linewidth]{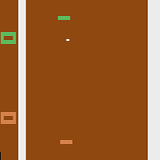}
  \captionof{figure}{\textbf{Pong rotated by 90 degrees}, \cite{atari}}
  \label{fig:test2}
\end{minipage}
\end{figure*}

\textbf{Experiment~1}: The source and target tasks were the same, i.e. game Pong (screenshot is shown in Figure \ref{fig:test1}). The main aim of this experiment was to verify the ability of the proposed approach to find the identity transformation.

\textbf{Experiment~2}: The source task was the original Pong while the target task was rotated Pong (see screenshot in Figure \ref{fig:test2}). The game remained the same, but all image frames were rotated by 90 degrees.

Each experiment consists of the following steps:
\begin{enumerate}[\hspace{0.5cm} 1)]
\item The agent played the \emph{source task} (standard Pong), learned the optimal policy by DQN and obtained the optimal $Q$-function $Q_S$, environment model $F$ and experience memory $\textbf{M}_S$ containing $10 000$ data entries collected at the end of the game.
\item The agent played the \emph{target task} (standard Pong in Experiment~1 or rotated Pong in Experiment~2) using random policy and obtained data for experience memory $\textbf{M}_T$ containing $10 000$ data entries.
\item The agent started learning the correspondence function $\mathcal{C}$ using the method from Section \ref{ch:tl learning} with the $Q$-function $Q_S$, environment model $F$ and experience memories $\textbf{M}_S$ and $\textbf{M}_T$,
\item For every $1000$ learning steps, the agent:
\begin{itemize}
\item suspends learning of correspondence function $\mathcal{C}$,
\item uses learnt $\mathcal{C}$ and the $Q$-function transformed from the \emph{source task}, see (\ref{e: transformed Q}), to play five games of the \emph{target task}, and 
\item computes the average accumulated reward per game.
\end{itemize}
\item The agent played the \emph{target task} while using the learned correspondence\footnote{the correspondence function that achieved the highest average accumulated reward per game in the previous step was used here} and  $Q$-function $Q_S$ transferred from the \emph{source} task. At the same time the agent uses DQN and fixed $\mathcal{C}$ to continuously fine-tune $Q$-function $Q_T$ of the target task.
\end{enumerate}
The \emph{key metric} to evaluate the success of the knowledge transfer was the average accumulated reward per game.

\noindent\textbf{Baseline methods:} The results are compared with two baselines—using \ac{GAN} and \ac{CycleGAN} methods \cite{gan}, \cite{cyclegan}, which have recently been applied for knowledge transfer in similar settings \cite{Gamrian}. Experiment 2 also includes \emph{fine-tuning} the $Q$-function from the source task as a baseline, as it is a commonly used transfer learning method.

The following sections provide the key details of the experiments performed and their results. 

\subsection{Experiment 1}
\label{s:experiment1}
This experiment aimed to test transfer learning when \emph{source} and \emph{target} tasks are identical.

$G_S$ and $G_T$ generators (see Section \ref{s:correspondence learning}) were constructed as neural networks with convolutional layers. Their specific architecture was taken from \cite{generator_arch_resnet}. The discriminators $D_S$ and $D_T$ were also constructed as neural networks with convolutional layers with the architecture as in \cite{discriminator_arch_resnet}.

The parameters of all of the networks were initialized from Gaussian distribution $N(0, 0.02)$. The transfer learning with the loss (\ref{e:final loss}) was tested for all the combinations of the parameters:  $\lambda_{Cyc} \in \{0, 1, 10\}$, $\lambda_Q \in \{0, 1\}$ and $\lambda_M \in \{0, 1, 10\}$.

\begin{figure*}[!t]
\centering
\begin{minipage}{.49\textwidth}
  \centering
  \includegraphics[width=.9\linewidth]{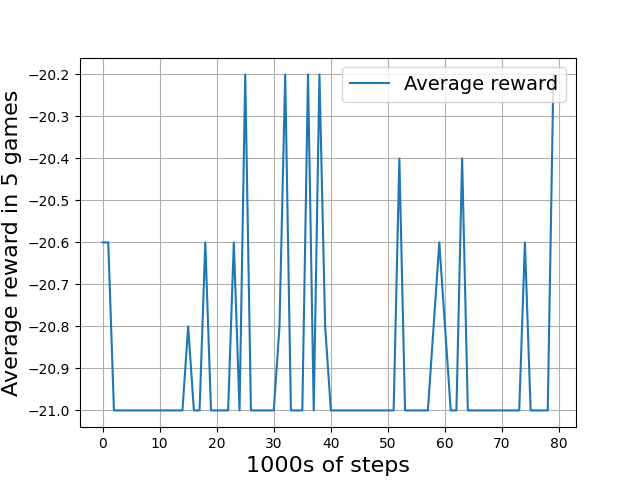}
  \newline
  a) \textbf{GAN} \newline $\lambda_{Cyc} = \lambda_Q = \lambda_M = 0$
\end{minipage}
\begin{minipage}{.49\textwidth}
  \centering
  \includegraphics[width=.9\linewidth]{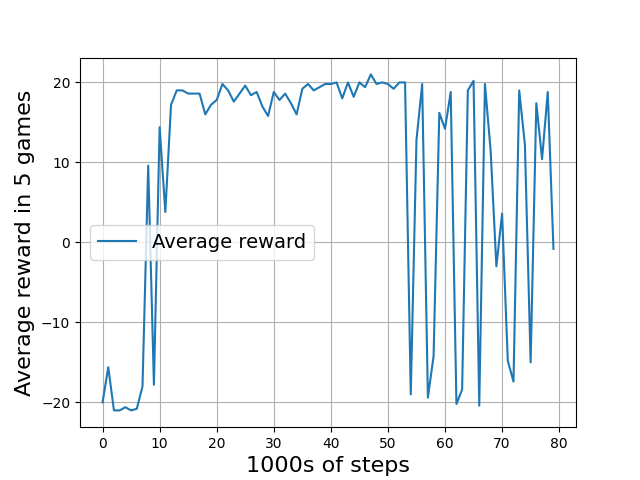}
  \newline
  b) \textbf{CycleGAN} \newline $\lambda_{Cyc} = 10,  \lambda_Q = \lambda_M = 0$
\end{minipage}
\vspace{0.7cm}

\begin{minipage}{.48\textwidth}
  \centering
  \includegraphics[width=.9\linewidth]{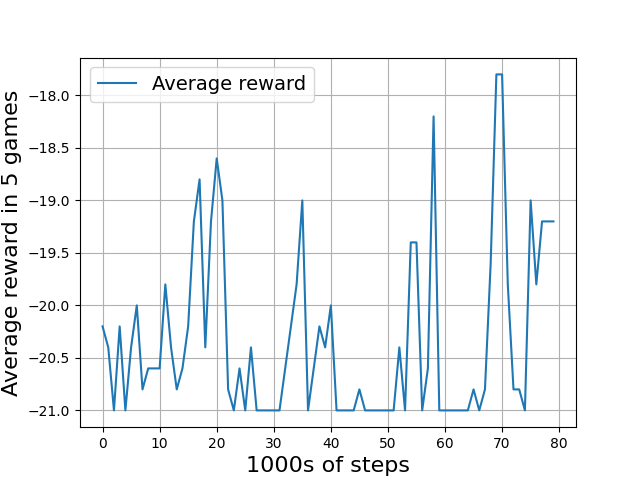}
  \newline
  c) Loss (\ref{e:final loss}) with $\lambda_{Cyc} = 0$,  $\lambda_Q = 1$, $\lambda_M = 0$
\end{minipage}
\begin{minipage}{.49\textwidth}
  \centering
  \includegraphics[width=.9\linewidth]{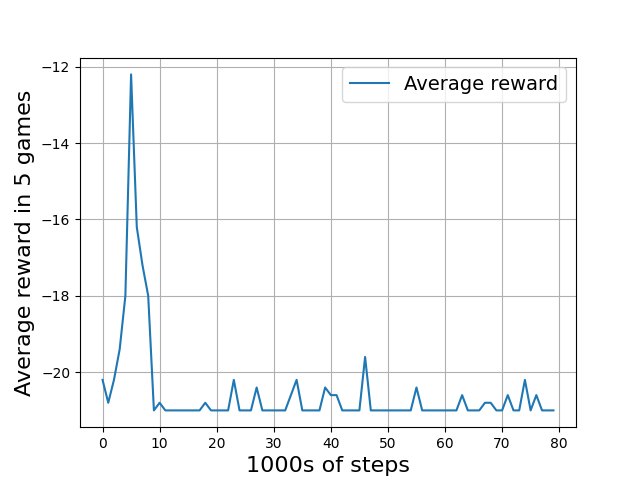}
  \newline
  d) Loss (\ref{e:final loss}) with $\lambda_{Cyc} = 0$,  $\lambda_Q = 0$, $\lambda_M = 10$
\end{minipage}
\vspace{0.7cm}
\begin{minipage}{.49\textwidth}
  \centering
  \includegraphics[width=.9\linewidth]{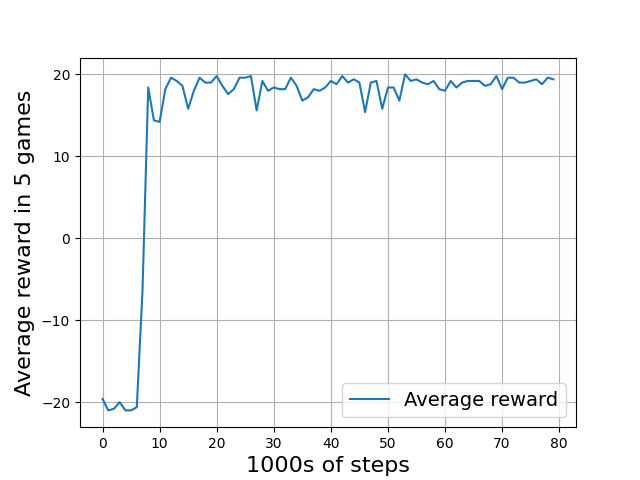}
  \newline
  e) Loss (\ref{e:final loss}) with $\lambda_{Cyc} = 1$,  $\lambda_Q = 1$, $\lambda_M = 1$
\end{minipage}
\caption{\label{f:exp1 average reward} \textbf{Experiment 1}: Average accumulated reward per game when playing five games with the transformed $Q$-function (\ref{e: transformed Q}). The agent paused the correspondence function learning each $1000$ learning steps and played five games where the average reward gained per game is displayed. The performance is shown for different values of loss parameters $\lambda_{Cyc}$, $\lambda_Q$ and $\lambda_M$. Figure \ref{f:exp1 average reward}a and \ref{f:exp1 average reward}b show the \textbf{baselines} using \emph{\ac{GAN}} and \emph{\ac{CycleGAN}} methods.}
 \vspace{0.5cm}
\end{figure*}

\subsubsection*{Results}

\begin{figure*}[!t]
\centering
\begin{minipage}{.99\textwidth}
  \centering
  \includegraphics[width=.99\linewidth]{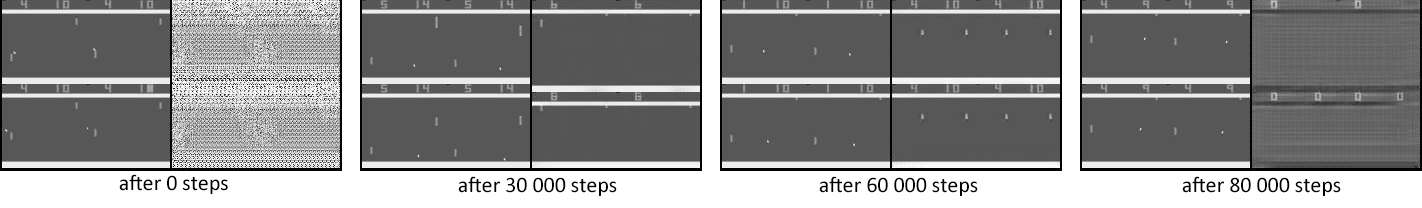}
  \newline
  a) \textbf{GAN} $\lambda_{Cyc} = 0$, $\lambda_Q = 0$, $\lambda_M = 0$
\end{minipage}

\vspace{0.5cm}

\begin{minipage}{.99\textwidth}
  \centering
  \includegraphics[width=.99\linewidth]{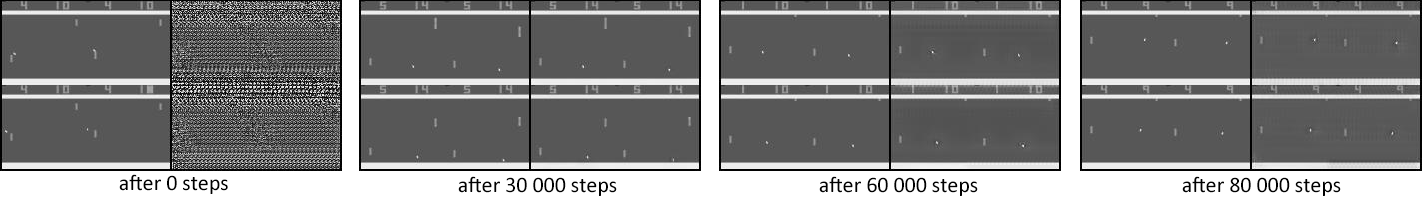}
  \newline
  b) \textbf{CycleGAN} $\lambda_{Cyc} = 10,  \lambda_Q = 0$, $\lambda_M = 0$
\end{minipage}

\vspace{0.5cm}

\begin{minipage}{.99\textwidth}
  \centering
  \includegraphics[width=.99\linewidth]{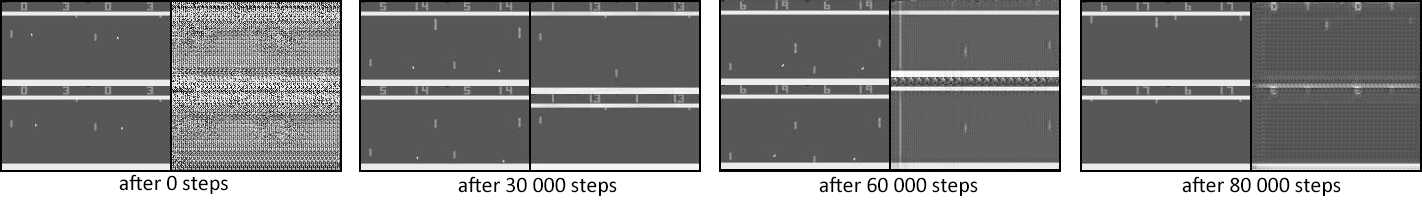}
  \newline
  c) $\lambda_{Cyc} = 0$,  $\lambda_Q = 1$, $\lambda_M = 0$
\end{minipage}

\vspace{0.5cm}

\begin{minipage}{.99\textwidth}
  \centering
  \includegraphics[width=.99\linewidth]{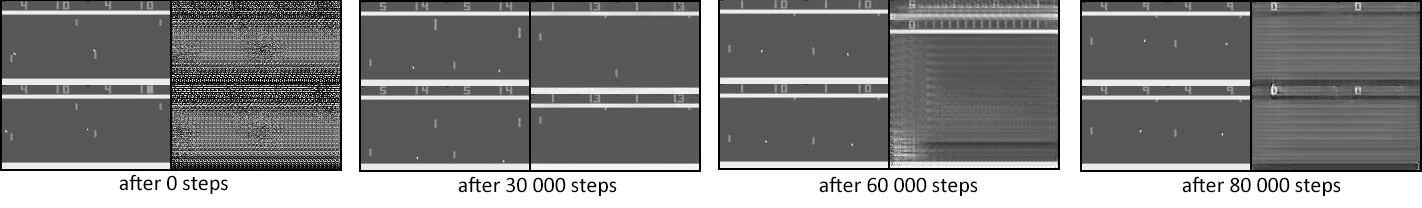}
  \newline
  d) $\lambda_{Cyc} = 0,  \lambda_Q = 0$, $\lambda_M = 10$
\end{minipage}

\vspace{0.5cm}

\begin{minipage}{.99\textwidth}
  \centering
  \includegraphics[width=.99\linewidth]{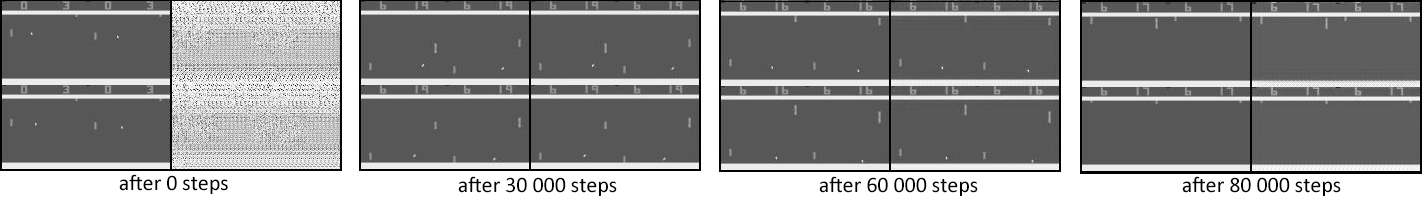}
  \newline
  e) $\lambda_{Cyc} = 1,  \lambda_Q = 1$, $\lambda_M = 1$
\end{minipage}
\caption{\label{f:exp1 progress}\textbf{Experiment~1}: Screenshots of the game depicting the progress of learning correspondence function $\mathcal{C}$, \eqref{e: specific correspondence function}, after $0, 30000, 60000$ and $80000$ steps. The results are shown for different values of parameters $\lambda_{Cyc}$, $\lambda_Q$ and $\lambda_M$ (\ref{e:final loss}). The left parts are game frames of the \emph{target} task  serving as states, and the right parts are the same states mapped by the learned correspondence function, $\mathcal{C}$.}
\end{figure*}

\begin{figure}[!t]
\centering
  \includegraphics[width=.5\linewidth]{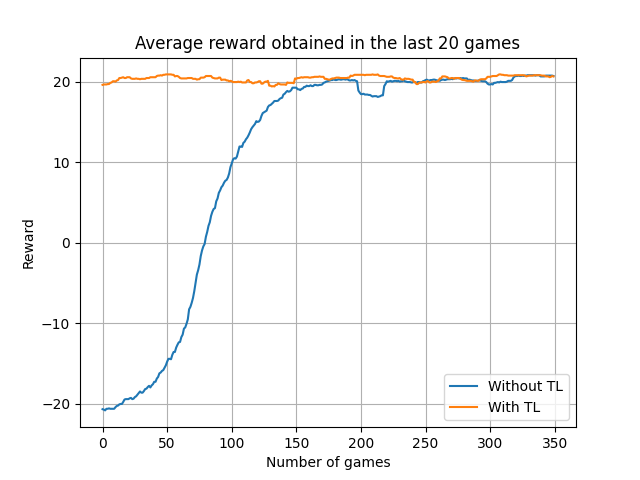}
  
  \caption{\label{f:exp1 main TL} Moving average of reward per game computed from the last $20$ games depending on the number of \textit{Pong} games played. The blue line denotes learning from scratch, i.~e.~without \ac{TL}. The orange line denotes the case with \ac{TL}, i.e. when the agent learns the correspondence function and uses the transformed $Q$-function (\ref{e: transformed Q}). The $Q$-function $Q_T$ is continuously learned during the game in both cases. }

\end{figure}

The results presented in Figure \ref{f:exp1 average reward} - Figure \ref{f:exp1 main TL} highlight the effectiveness of the proposed method in comparison to baseline models. After every 1000 learning steps, the agent pauses to play five games, and the average reward per game is recorded.

In Figure \ref{f:exp1 average reward}, the best results are observed when all loss components ($\Lagr_{Cyc}$, $\Lagr_Q$, and $\Lagr_M$) are included in the total loss function \eqref{e:final loss} with parameters $\lambda_{Cyc}=\lambda_Q=\lambda_M=1$, as shown in Figure \ref{f:exp1 average reward}e. This configuration achieves nearly the maximum reward (21), indicating that the method transfers knowledge effectively and optimizes performance.

The significance of the new components is demonstrated by Figure \ref{f:exp1 average reward}c and Figure\ref{f:exp1 average reward}d. When only one of the new components ($\Lagr_Q$ or $\Lagr_M$) is included, the performance drops noticeably. This result emphasizes that both components are critical to achieving successful knowledge transfer. Other parameter combinations did not yield meaningful results and are therefore not presented here.

In contrast, the baseline methods perform poorly. The \ac{GAN} baseline (Figure \ref{f:exp1 average reward}a) fails to yield meaningful results, while the \ac{CycleGAN} baseline (Figure \ref{f:exp1 average reward}b) shows initial success, but its performance quickly becomes unstable, indicating that it cannot maintain an effective correspondence function over time. The fluctuations observed in the CycleGAN curve in Figure \ref{f:exp1 average reward}b stem from the adversarial nature of the training process, and adding the proposed losses appears to help mitigate this instability.

Figure \ref{f:exp1 average reward} visually demonstrates the correspondence between the source and target tasks, confirming that the best performance is obtained when all components of the loss function are active. Although the \ac{CycleGAN} baseline shows some visual accuracy, its inconsistency is evident in the unstable reward progression.

Finally, Figure \ref{f:exp1 main TL} compares the performance of an agent learning from scratch with one that transfers knowledge using the proposed method. The agent that reuses previously learned knowledge reaches high performance almost immediately, while the agent learning from scratch requires much more time to reach the same level. This highlights the efficiency of our method both in transferring knowledge and reducing training time.

\subsection{Experiment 2}
\label{s:experiment2}

In Experiment~2, the \emph{target task} is the original Pong with image frames rotated by $90$ degrees (see Figure~\ref{fig:test2}). 

Generators $G_S$ and $G_T$, (see \eqref{e: generators} and Section~\ref{s:correspondence learning}) are constructed as neural networks. Two types of generators were used in the experiment. The architecture of the first one, referred to here as the \textbf{resnet generator}, was taken from \cite{generator_arch_resnet} and then followed by a rotation layer, see \cite{spatial-transformer}. The second type, referred to as the \textbf{rotation generator}, was composed of the mentioned rotation layer only. Discriminators $D_S$ and $D_T$ are constructed by neural networks with convolutional layers with the architecture as in \cite{discriminator_arch_resnet}.

\subsubsection*{Results}

\begin{figure*}[!t]
\centering
\begin{minipage}{.45\textwidth}
  \centering
  \includegraphics[width=.9\linewidth]{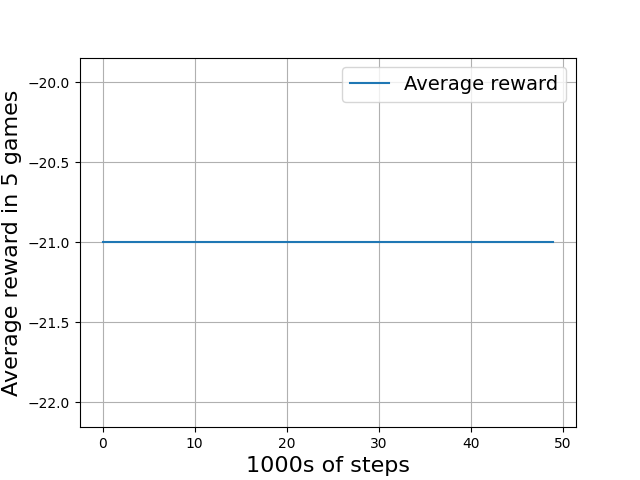}
  \newline
  a) \textbf{Rotation generator} \newline using \textbf{GAN}
  
  \vspace{0.5cm}
  
  \includegraphics[width=.9\linewidth]{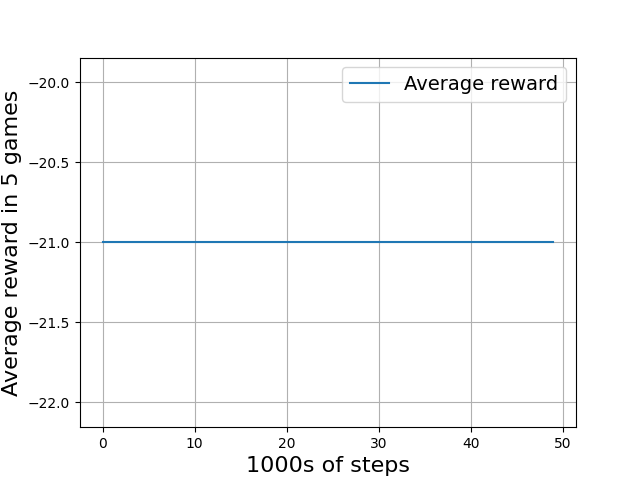}
  \newline
  c) \textbf{Rotation generator} \newline using \textbf{CycleGAN}

\vspace{0.5cm}
  
  \includegraphics[width=.9\linewidth]{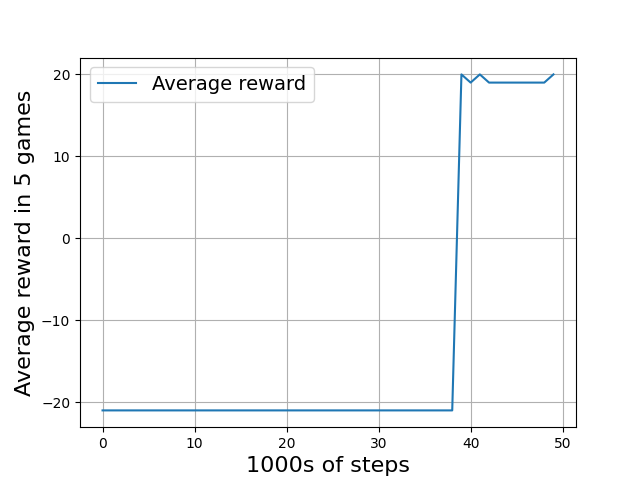}
  \newline
  e) \textbf{Rotation generator} \newline  with $\lambda_{Cyc} = \lambda_Q = 0$, $\lambda_M = 10$
\end{minipage}
\begin{minipage}{.45\textwidth}
  \centering
  \includegraphics[width=.9\linewidth]{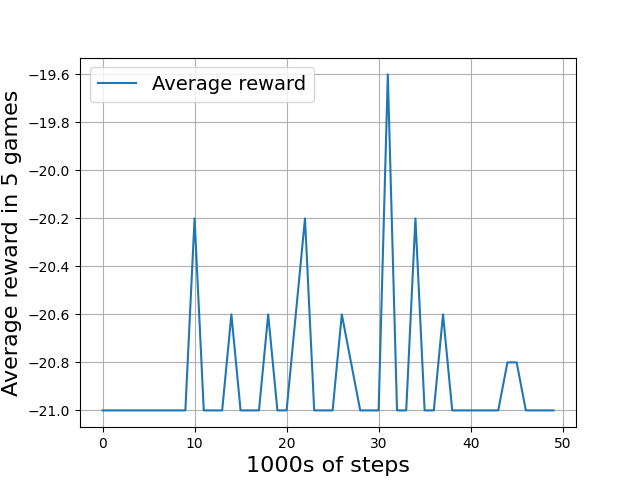}
  \newline
  b) \textbf{Resnet generator} \newline using \textbf{GAN}
  
\vspace{0.5cm}  
  
 \includegraphics[width=.9\linewidth]{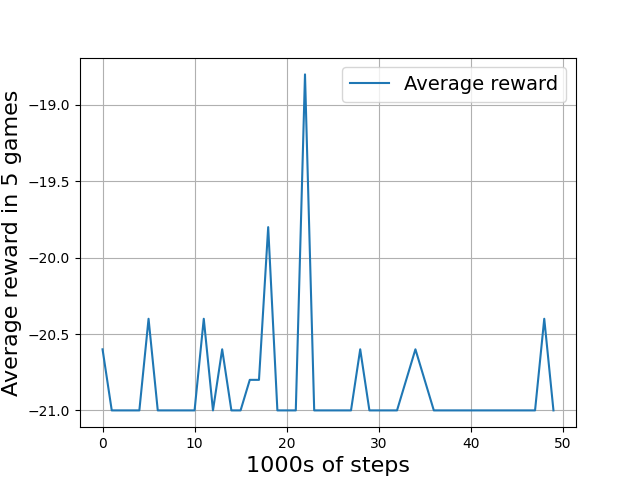}
  \newline
  d) \textbf{Resnet generator} \newline using \textbf{CycleGAN}
  
  \vspace{0.5cm}
  
   \includegraphics[width=.9\linewidth]{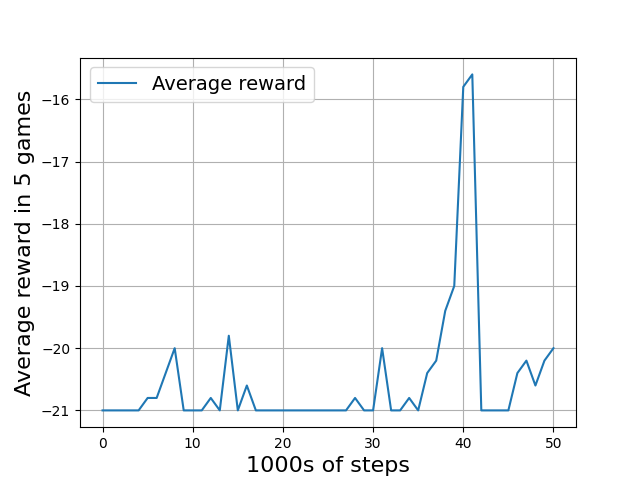}
  \newline
  f) \textbf{Resnet generator} \newline with $\lambda_{Cyc} = 1,  \lambda_Q = 0$, $\lambda_M = 1$  
  
\end{minipage}
\vspace{0.7cm}
\caption{\label{f:exp2 average reward} Average accumulated reward in five games when playing Rotated Pong with the transformed $Q$-function (\ref{e: transformed Q}). The agent paused the correspondence function learning each $1000$ learning steps and played five games where the average reward gained per game is displayed. The results are shown for the \textbf{rotation} and the \textbf{resnet} generator with the best settings of the loss parameters in each case (e, f) as well as with using \textbf{GAN} and \textbf{CycleGAN} baselines (a-d).
}
 \vspace{1cm}
\end{figure*}

\begin{figure*}[!t]
\centering
\begin{minipage}{.99\textwidth}
  \centering

  \includegraphics[width=.99\linewidth]{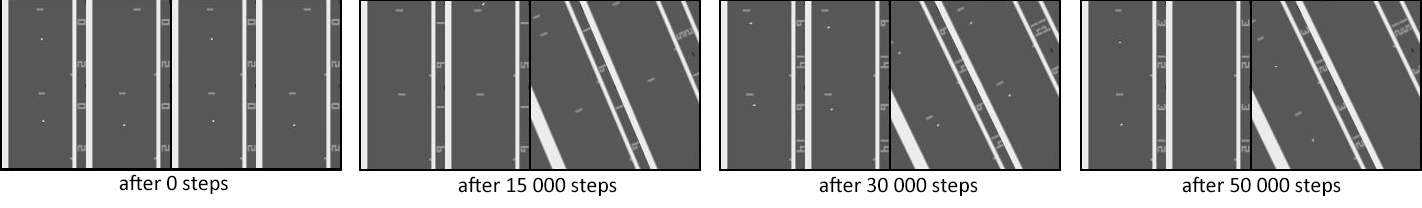}
  \newline
  a) Rotation generator using \textbf{GAN}

\vspace{0.5cm}
  
  \includegraphics[width=.99\linewidth]{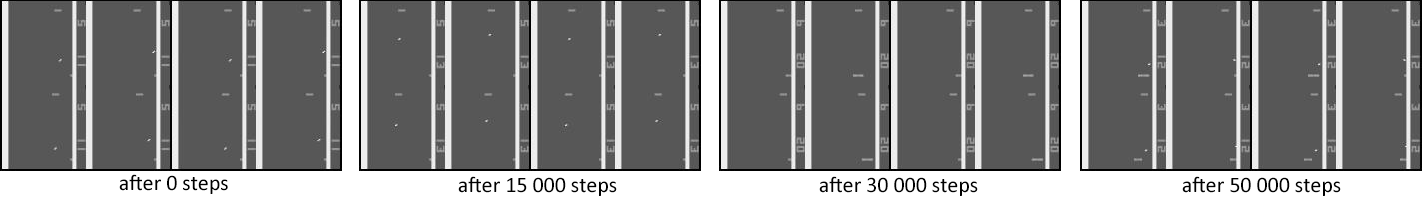}
  \newline
  b) Rotation generator using \textbf{CycleGAN}

\vspace{0.5cm}  
  
  \includegraphics[width=.99\linewidth]{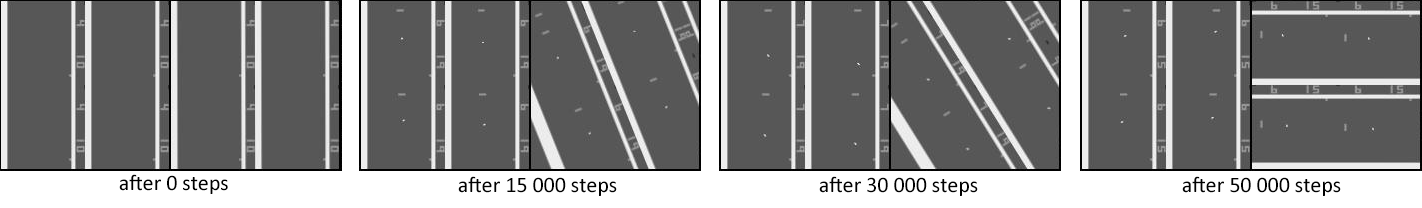}
  \newline
  c) Rotation generator with $\lambda_{Cyc} = 0$, $\lambda_Q = 0$, $\lambda_M = 10$
\end{minipage}

\vspace{0.5cm}

\begin{minipage}{.99\textwidth}
  \centering
  
\includegraphics[width=.99\linewidth]{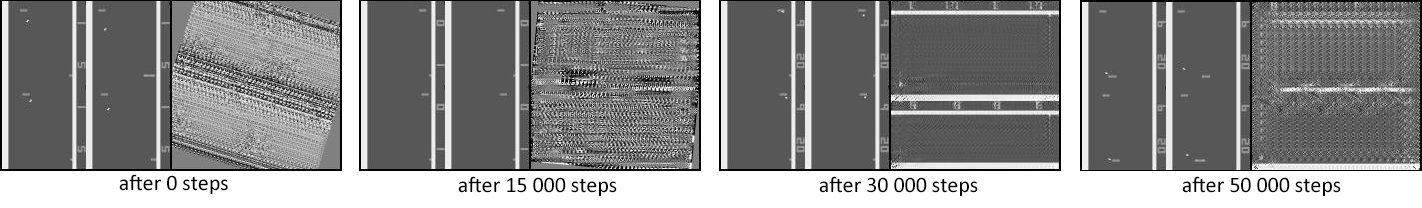}
  \newline
  d) Resnet generator using \textbf{GAN}

  \vspace{0.5cm}
  
\includegraphics[width=.99\linewidth]{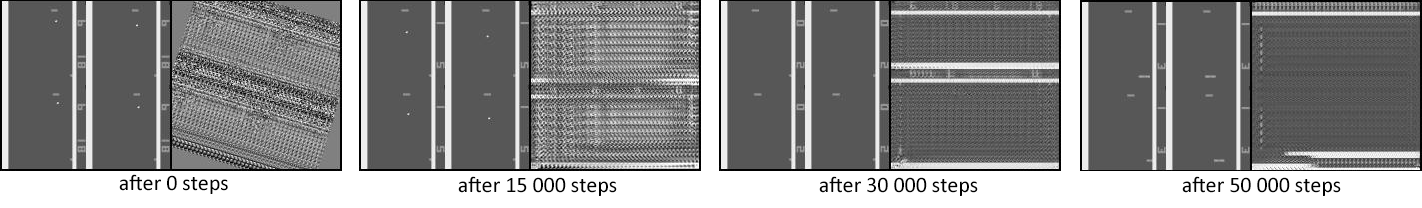}
  \newline
  e) Resnet generator using \textbf{CycleGAN}

  \vspace{0.5cm}  
  
  \includegraphics[width=.99\linewidth]{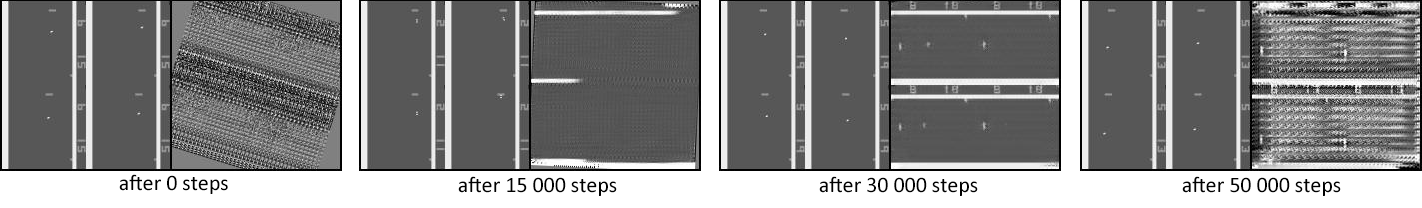}
  \newline
  f) Resnet generator with $\lambda_{Cyc} = 1,  \lambda_Q = 0$, $\lambda_M = 1$
\end{minipage}
\caption{\label{f:exp2 progress}
\textbf{Experiment 2}: Screenshots of the game depicting progress in learning the correspondence function $\mathcal{C}$ (\ref{e: specific correspondence function}) after 0, 15000, 30000 and 50000 steps. The results are shown for the rotation and the resnet generators with the best settings of parameters $\lambda_{Cyc}$, $\lambda_Q$ and $\lambda_M$ (\ref{e:final loss}) as well as with using \textbf{GAN} and \textbf{CycleGAN} baselines. The left parts of the pictures are game frames of the \emph{target} task representing the states, and the right parts are the same states transformed by the correspondence function $\mathcal{C}$.
}
 \vspace{1cm}
\end{figure*}

\begin{figure}[!t]
\centering

  \centering
  \includegraphics[width=.9\linewidth]{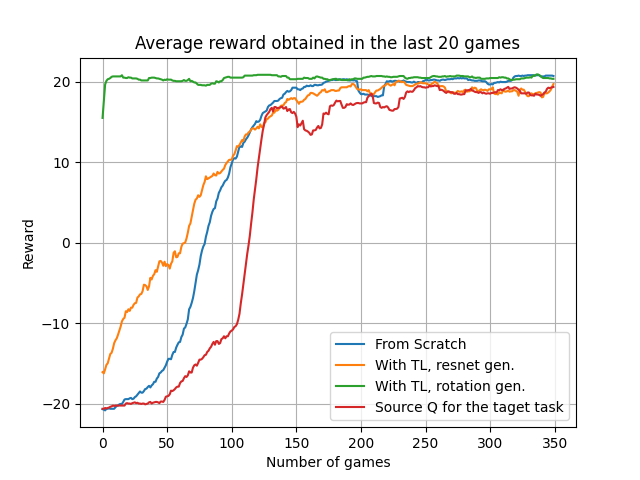}
  \caption{\label{f:exp2 main TL} Moving average of reward per game computed from the last 20 games depending on the number of played games for the game rotated Pong for four different agents - an agent learning the game from scratch (blue line), an agent using the correspondence function learned with the resnet generator (orange line), an agent using the correspondence function learned with the rotation generator (green line) and an agent reusing only the $Q$-function without any correspondence function (red line). The agents were continuously learning the $Q$-function.  
  }

\end{figure}

The proposed approach was tested with various values of the loss parameters $\lambda_{Cyc}$, $\lambda_Q$, and $\lambda_M$ (from \eqref{e:final loss}). Figure \ref{f:exp2 average reward} - Figure  \ref{f:exp2 main TL} present the best-achieved performance of our method, compared to baseline methods.

Figure \ref{f:exp2 average reward} depicts the average reward per game over five games. Similar to Experiment 1, after every 1000 learning steps, the agent pauses the learning of the correspondence function and plays five games of the target task. The results show that the \emph{rotation} generator achieves nearly perfect knowledge transfer, with rewards approaching the maximum score of 21. This indicates that the \emph{rotation} generator can establish an effective correspondence between the source and target tasks, enabling high-performance transfer.

In contrast, although the \emph{resnet} generator did not yield perfect results, it still learned a reasonable correspondence function, particularly after 50,000 steps, as shown in Figure \ref{f:exp2 progress}. This partially successful correspondence was then used to fine-tune the $Q$-function for the target task. Importantly, this fine-tuning process led to much better results than training the $Q$-function from scratch, demonstrating that even an imperfect correspondence can significantly accelerate learning.
G
The baseline methods, using standard \ac{GAN} and \ac{CycleGAN}, failed to produce any usable correspondence for knowledge transfer. This is clearly illustrated in Figure~\ref{f:exp2 average reward}a and Figure~\ref{f:exp2 average reward}b, where the agent's poor performance highlights the limitations of these methods for reinforcement learning tasks.

Figure \ref{f:exp2 progress}a illustrates the progression of the correspondence function learned by the \emph{rotation} generator, which consistently mapped the source task to the target task correctly. In contrast, Figure~\ref{f:exp2 progress}b indicates the slower, but ultimately reasonable, progress made by the \emph{resnet} generator, further emphasizing the benefits of the \emph{rotation} generator in establishing task similarity.

Lastly, Figure~\ref{f:exp2 main TL} compares the performance of the agent in the rotated Pong task under different conditions:
\begin{enumerate}
    \item When learning from scratch, performance improves slowly over time.
    \item When using the correspondence function learned by the \emph{resnet} generator, performance improves much faster at the start.
    \item When using the \emph{rotation} generator, the agent achieves immediate reuse of prior knowledge and performs at a high level from the beginning.
    \item When fine-tuning the $Q$-function from the source task without considering correspondence, performance was worse than learning from scratch, likely due to overfitting to the source task.
\end{enumerate}

This comparison highlights the advantages of the proposed method, which enables seamless knowledge transfer, significantly reduces training time, and improves initial performance. In contrast, fine-tuning the $Q$-function (used as one of the baselines) without proper alignment between tasks leads to poor results, underscoring the importance of task-specific correspondence functions.

\section{Conclusion and Discussion}
\label{ch:conclusion}

This paper presented a novel method for efficient one-to-one knowledge transfer between reinforcement learning tasks. Our approach modifies \ac{CycleGAN} specifically for reinforcement learning by incorporating a new loss function that includes the $Q$-function and environment model from the source task. Through experiments on the 2-D Atari game Pong, we demonstrated that our method outperforms baseline models such as GAN and CycleGAN, providing faster learning and better performance, particularly in scenarios where task environments differ.

One of the key findings of this work is the importance of the network architecture when learning the correspondence function. While both the rotation-based and convolutional generators achieved reasonable results, the rotation-based generator yielded superior performance. This suggests that convolutional layers, commonly used in image-based tasks, may not be optimal for reinforcement learning transfer learning tasks. Future research should explore other architectures, such as transformers, \cite{transformers}, which may further improve generalisation.

In comparison to other knowledge transfer methods, our approach has the advantage of being applicable to a variety of domains without the need for paired data, allowing it to handle diverse \ac{RL} tasks with varying state formats. However, we acknowledge some limitations. The current method struggles with tasks that have low similarity, and we have not yet explored transferring knowledge from multiple source tasks or automatically selecting the most relevant source task.

\subsubsection*{Future Directions}
\begin{enumerate}
 \item \textbf{Expanding the validation}: testing the proposed method on a broader range of RL tasks to assess its generalization ability and robustness.
	\item \textbf{Knowledge transfer in low-similarity tasks}: investigating how to transfer knowledge between tasks with low similarity.
    \item \textbf{Identifying relevant knowledge}: exploring methods to identify and transfer relevant knowledge from multiple source tasks.
    \item \textbf{Source task selection}: developing strategies for selecting the most relevant source tasks for transfer.
    \item \textbf{Alternative network architectures}: researching alternative network architectures, like transformers, to enhance correspondence learning.
\end{enumerate}

In conclusion, our approach represents a significant advancement toward practical and flexible knowledge transfer in reinforcement learning. However, several challenges remain that future work must address to enhance the robustness and adaptability of such systems.

\textbf{Method implementation}: The method implementation in Python is available at \emph{https://github.com/marko-ruman/RL-Correspondence-Learner}

\textbf{Data availability statement}: The datasets generated and/or analysed during the current study are available from the corresponding author on reasonable request.

\textbf{Ethical approval}: This article does not contain any studies with human participants performed by any of the authors.

\appendix[]\label{appendix}
\section*{DQN algorithm}
Algorithm~\ref{a:dqn} summarises the DQN algorithm used in the text (see Section \ref{s:dqn} for details). $\theta$ denotes parameters of the source network and $\theta^T$ are parameters of the target network. Both networks have the same architecture.
\begin{algorithm}[!t]
	\caption{DQN \label{a:dqn}}
	\begin{algorithmic}[1]
		\REQUIRE{initial parameters $\theta$ of $Q$-function $Q(s,a,\theta)$, learning rate $\alpha \in (0,1)$, discount factor $\gamma \in (0,1)$, exploration rate $\epsilon \in (0,1)$, size of the experience memory $n_M$, size of the learning mini-batch $n_B$, number of steps for target network synchronization $n_U$}
		\STATE Initialize experience memory size $n_M$
		\STATE Set parameters of the target network $\theta^T = \theta$
		\FOR{$t=1,2,...,$ till convergence}
		\STATE With exploration $\epsilon$ perform random action $a_t$
		otherwise select $a_t = \underset{\is{a}{a}}{\mathrm{argmax}}\hspace{0.07cm}Q(s,a \mid  \theta)$
		\STATE Get next state $s_{t+1}$ and reward $r_t$
		\STATE If the memory is full, remove the oldest data record 
		\STATE Store $(s_t, a_t, r_t, s_{t+1})$ in experience memory $\textbf{M}$
		\STATE Sample a random mini-batch of size $n_B$ $(s_j, a_j, r_j, s_{j+1})_{j \in Rand(n_B)} \in \textbf{M}$
		\FOR{every $j$}
		\IF{$s_{j+1}$ is a terminal state}
		\STATE $\text{target}_j = r_j$
		\ELSE
		\STATE $\text{target}_j  = r_j + \gamma \hspace{0.07cm}\underset{\is{a'}{a}}{\mathrm{max}}\hspace{0.07cm}Q(s_{j+1},a' \mid \theta^T)$
		\ENDIF
		\ENDFOR
		
		\STATE Perform a gradient descent on $\left( \left(\text{target}_j - Q(s_j,a_j \mid  \theta)\right)^2\right)_{j \in Rand(n_B)}$ with Huber loss, \cite{huber}, with respect to parameters $\theta$
		\STATE Every $n_U$ steps set $\theta^T = \theta$
		\ENDFOR
		\ENSURE{$Q$-function $Q(s,a)$, experience memory $\textbf{M}$}
	\end{algorithmic}
\end{algorithm}

\section*{Implementation details}
\paragraph{\ac{CycleGAN} generator architectures} The architectures of generators $G_S$ and $G_T$ in Experiment 1 (Section \ref{s:experiment1}) and the resnet generators $G_S$ and $G_T$ in Experiment 2 (Section \ref{s:experiment2}) were taken from \cite{cyclegan}. The 9 residual blocks version was used. Below, we follow
the naming convention used in \cite{cyclegan}.

Let \texttt{c7s1-f} denote a $7 \times 7$ Convolution-BatchNorm-ReLU layer with $f$ filters and stride 1. \texttt{df} denotes a $3 \times 3$
Convolution-BatchNorm-ReLU layer with $f$ filters and
stride 2. Reflection padding was used to reduce artefacts.
\texttt{Rf} denotes a residual block that contains two $3 \times 3$ convolutional layers with the same number of filters ($f$) on both
layers. \texttt{uf} denotes a $3 \times 3$ fractional-strided-Convolution-
BatchNorm-ReLU layer with $f$ filters and stride 2.

The network architecture consisted of:

\texttt{c7s1-64,d128,d256,R256,R256,R256,R256,
R256,R256,R256,R256,R256,u128,u64,c7s1-3}

The \emph{rotation} generator contained just one rotation layer, see \cite{spatial-transformer}.

\paragraph{Discriminator architectures} For discriminator networks $D_S$ and $D_T$ in all the experiments, $70 \times 70$ PatchGAN was used, see \cite{discriminator_arch_resnet}. Let \texttt{Cf} denote a
$4 \times 4$ Convolution-BatchNorm-LeakyReLU layer with $f$
filters and stride 2. After the last layer, a convolution to produce a 1-dimensional output was used. Leaky ReLUs were used with a slope of 0.2. 

The discriminator architecture was:

\texttt{C64,C128,C256,C512}.

\paragraph{$Q$-function architecture} $Q$-function had architecture taken from \cite{dqn}. Let \texttt{c-k-s-f} denote a $k \times k$ Convolution-ReLU layer with stride $s$ and $f$ filters and \texttt{f-o} is a Fully connected-ReLU layer with $o$ outputs. The $Q$-function architecture was:

\texttt{c-8-1-32,c-4-2-64,c-3-1-64,f-512,f-6}.

\paragraph{Environment model architecture} The environment model $F$ had the same architecture as the generators $G_S$ and $G_T$ with one difference: the fifth residual block received one-hot encoded actions as an additional input.

The architecture of the environment model was then as follows:

\texttt{c7s1-64,d128,d256,R256,R256,R256,R256,
R262,R262,R262,R262,R262,u128,u64,c7s1-3}.

\section*{Training}

All the networks are trained from scratch with weights initialized from a Gaussian distribution $N(0, 0.02)$. 

The environment model, $F$, was trained with Adam optimizer, \cite{adam}, with the learning rate of 0.001, batch size of 16 and it was trained for 50 epochs. 

For the training of the $Q$-function, RMSprop optimiser, \cite{rmsprop}, was used. The learning rate was 0.0001, and the batch size was 32. The other parameters of $Q$-learning were identical to those in \cite{dqn}.

Generators $G_S$ and $G_T$ and discriminators $D_S$ and $D_T$ were jointly trained using Adam optimizer with an initial learning rate of 0.0002 which was linearly decayed to zero. The training took four epochs.

\bibliographystyle{unsrt}

\bibliography{knowledge_transfer_in_deep_rl.bib}

\end{document}